\pdfoutput=1

\documentclass[11pt]{article}

\usepackage[]{acl}

\usepackage{times}
\usepackage{latexsym}

\usepackage[T1]{fontenc}

\usepackage[utf8]{inputenc}

\usepackage{microtype}

\usepackage{algorithm}
\usepackage{algorithmic}
\usepackage{graphicx} 
\usepackage{xcolor}
\usepackage{amsfonts}
\usepackage{booktabs}
\usepackage{amsmath}
\usepackage{bm}
\usepackage{multirow}
\usepackage{multicol}
\usepackage{natbib}
\usepackage{bibentry}
\usepackage{soul}
\usepackage{flushend}
\usepackage{xspace}
\usepackage{longtable}

\DeclareMathOperator*{\argmin}{arg\,min}
\DeclareMathOperator*{\elu}{ELU}
\DeclareMathOperator{\Tr}{Tr}
\newcommand{\nertag}[1]{\texttt{#1}}
\newcommand{\tokentag}[2]{$\textcolor{blue}{\text{#1}}_{\textcolor{green}{#2}}$}

\newcommand{\ours}{\textsc{CONTaiNER}\xspace}
\title{\ours: Few-Shot Named Entity Recognition via Contrastive Learning}

\author{Sarkar Snigdha Sarathi Das, Arzoo Katiyar, Rebecca J. Passonneau, Rui Zhang \\
        Pennsylvania State University\\
        \texttt{\{sfd5525, arzoo, rjp49, rmz5227\}@psu.edu} }

\begin{document}
\maketitle
\begin{abstract}
Named Entity Recognition (NER) in Few-Shot setting is imperative for entity tagging in low resource domains. Existing approaches only learn \emph{class-specific} semantic features and intermediate representations from source domains. This affects generalizability to unseen target domains, resulting in suboptimal performances. 
To this end, we present \ours, a novel contrastive learning technique that optimizes the inter-token distribution distance for Few-Shot NER. Instead of optimizing class-specific attributes, \ours optimizes a generalized objective of differentiating between token categories based on their Gaussian-distributed embeddings.
This effectively alleviates overfitting issues originating from training domains. 
Our experiments in several traditional test domains (OntoNotes, CoNLL'03, WNUT '17, GUM) and a new large scale Few-Shot NER dataset (Few-NERD) demonstrate that, {on average}, \ours outperforms previous methods by 3\%-13\% absolute F1 points while showing consistent performance trends, even in challenging scenarios where previous approaches could not achieve appreciable performance. The source code of \ours will be available at:  \url{https://github.com/psunlpgroup/CONTaiNER}. 
\end{abstract}

\section{Introduction}

Named Entity Recognition (NER) is a fundamental NLU task that recognizes mention spans in unstructured text and categorizes them into a pre-defined set of entity classes. In spite of its challenging nature, recent deep-learning based approaches~\cite{huang2015bidirectional,ma2016end,lample2016neural,peters2018deep,devlin2018bert} have achieved impressive performance.
As these supervised NER models require large-scale human-annotated datasets, few-shot techniques that can effectively perform NER in resource constraint settings have recently garnered a lot of attention.

\begin{figure}[t!]
\centering
\includegraphics[width=0.48\textwidth,height=0.09\textheight]{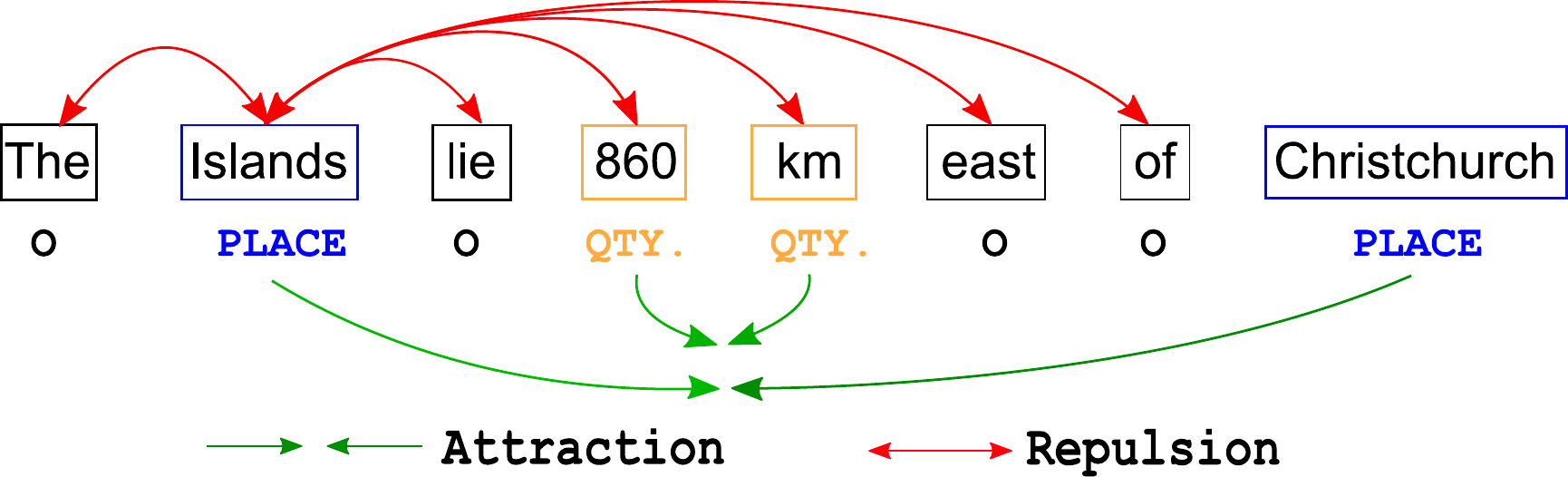}
\caption{Contrastive learning dynamics of a token (\emph{Islands}) with all other tokens in an example sentence from GUM \cite{Zeldes2017}. \ours decreases the embedding distance between tokens of the same category (\textcolor{blue}{\nertag{PLACE}}) while increasing the distance between different categories (
\textcolor[HTML]{FF9900}{QTY.} and \textcolor{black}{O}).}
\label{fig:fig_demonstration}
\vspace{-6mm}
\end{figure}

Few-shot learning involves learning unseen classes from very few labeled examples~\cite{fei2006one,lake2011one,bao2019few}. To avoid overfitting with the limited available data, meta-learning has been introduced to focus on \textit{how to learn}~\cite{vinyals2016matching, bao2019few}. \citet{snell2017prototypical} proposed Prototypical Networks to learn a metric space where the examples of a specific unknown class cluster around a single prototype. Although it was primarily deployed in computer vision, \citet{fritzler2019few} and \citet{hou2020few} also used Prototypical Networks for few-shot NER. \citet{yang2020simple}, on the other hand, proposed a supervised NER model that learns class-specific features and extends the intermediate representations to unseen domains. Additionally, they employed a Viterbi decoding variant of their model as "StructShot".

Few-shot NER poses some unique challenges that make it significantly more difficult than other few-shot learning tasks. First, as a sequence labeling task, NER requires label assignment according to the concordant context as well as the dependencies within the labels~\cite{lample2016neural, yang2020simple}. Second, in NER, tokens that do not refer to any defined set of entities are labeled as \nertag{Outside} (\nertag{O}). Consequently, a token that is labeled as \nertag{O} in training entity set may correspond to a valid target entity in test set. For prototypical networks, this challenges the notion of entity examples being clustered around a single prototype. As for Nearest Neighbor based methods such as~\citet{yang2020simple}, they are initially ``pretrained" with the objective of source class-specific supervision. As a result, the trained weights will be closely tied to the source classes and the network will project training set \nertag{O-tokens} so that they get clustered in embedding space. This will force the embeddings to drop a lot of useful features pertaining to its true target entity in the test set. Third, in few-shot setting, there are not enough samples from which we can select a validation set. This reduces the capability of hyperparameter tuning, which particularly affects template based methods where prompt selection is crucial for good performance \cite{cui2021template}. In fact, the absence of held-out validation set puts a lot of earlier few-shot works into question whether their strategy is truly "Few-Shot"~\cite{perez2021true}.

To deal with these challenges, we present a novel approach , \ours that harnesses the power of contrastive learning to solve Few-Shot NER. \ours tries to decrease the distance of token embeddings of similar entities while increasing it for dissimilar ones (Figure \ref{fig:fig_demonstration}). This enables \ours to better capture the label dependencies. Also, since \ours is trained with a generalized objective, it can effectively avoid the pitfalls of \nertag{O-tokens} that the prior methods struggle with. Lastly, \ours does not require any dataset specific prompt or hyperparameter tuning. Standard settings used in prior works \cite{yang2020simple} works well across different domains in different evaluation settings. 

Unlike traditional contrastive learners~\cite{chen2020simple, khosla2020supervised} that optimize similarity objective between point embeddings, \ours optimizes distributional divergence effectively modeling Gaussian Embeddings. While point embedding simply optimizes sample distances, Gaussian Embedding faces an additional constraint of maintaining class distribution through the variance estimation. Thus Gaussian Embedding explicitly models entity class distributions which not only promotes generalized feature representation but also helps in few-sample target domain adaptation. Previous works in Gaussian Embedding has also shown that mapping to a density captures representation uncertainties \cite{vilnis2014word} and expresses natural asymmetries \cite{qian2021conceptualized}  while showing better generalization requiring less data to achieve optimal performance \cite{bojchevski2017deep}.  
Inspired by these unique qualities of Gaussian Embedding, in this work we leverage Gaussian Embedding in contrastive learning for Few-Shot NER.

A nearest neighbor classification scheme during evaluation reveals that on average, \ours significantly outperforms previous SOTA approaches in a wide range of tests by up to 13\% absolute F1-points. In particular, we extensively test our model in both in-domain and out-of-domain experiments as proposed in~\citet{yang2020simple} in various datasets (CoNLL '03, OntoNotes 5.0, WNUT '17, I2B2). We also test our model in a large dataset recently proposed for Few-Shot NER - Few-NERD~\cite{ding2021few} where \ours outperforms all other SOTA approaches setting a new benchmark result in the leaderboard.

In summary, our contributions are as follows: (1) We propose a novel Few-Shot NER approach \ours that leverages contrastive learning to infer distributional distance of their Gaussian Embeddings. To the best of our knowledge we are the first to leverage Gaussian Embedding in contrastive learning for Named Entity Recognition.
(2) We demonstrate that \ours representations are better suited for adaptation to unseen novel classes, even with a low number of support samples.
(3) We extensively test \ours in a wide range of experiments using several datasets and evaluation schemes. In almost every case, our model largely outperforms present SOTAs establishing new benchmark results.

\begin{figure*}[ht!]
\centering
\includegraphics[width=0.97\textwidth,height=0.27\textheight]{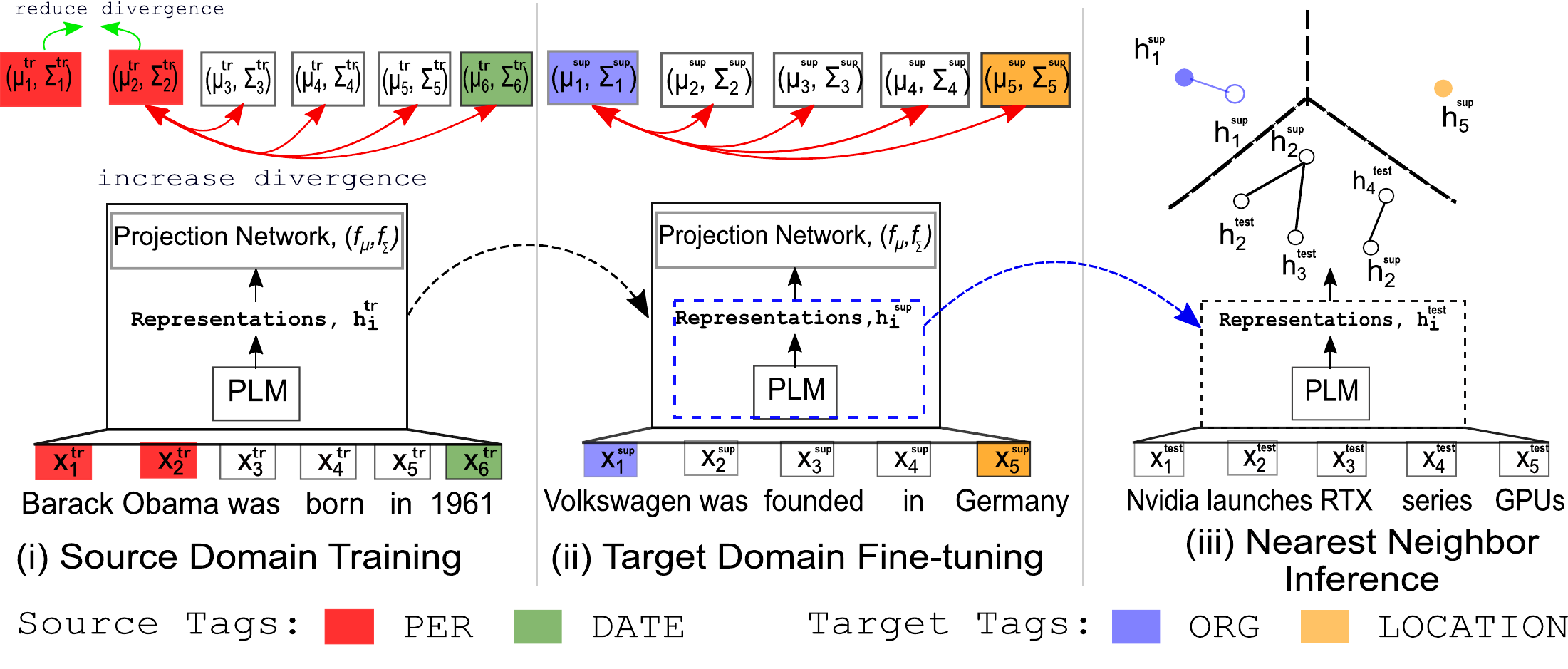}
\caption{Illustration of our proposed \ours framework based on Contrastive Learning over Gaussian Embedddings: (i) Training in source domains using training NER labels \nertag{PER} and \nertag{DATE}, (ii) Fine-tuning to target domains using target NER labels \nertag{ORG} and \nertag{LOCATION}, (iii) Assigning labels to test samples via Nearest Neighbor support set labels.}
\label{fig:fig_architecture}
\end{figure*}
\section{Task Formulation}
 
Given a sequence of $n$ tokens $\{x_1, x_2, \ldots x_n\}$, NER aims to assign each token $x_i$ to its corresponding tag label $y_i$.

\paragraph{Few-shot Setting} For Few-shot NER, a model is trained in a source domain with a tag-set $\{C_{(i)}^s\}$ and tested in a data-scarce target domain with a tag-set $\{C_{(j)}^{d}\}$ where $i,j$ are index of different tags. Since $\{C_{(i)}^{s}\} \cap \{C_{(j)}^{d}\} = \emptyset$, it is very challenging for models to generalize to unseen test tags. In an \emph{N-way K-shot} setting, there are $N$ tags in the target domain $|\{C_{(j)}^{d}\}| = N$, and each tag is associated with a support set with \emph{K} examples. 

\paragraph{Tagging Scheme} 
For fair comparison of \ours against previous SOTA models, we follow an IO tagging scheme where \nertag{I-type} represents that all of the tokens are inside an entity, and \nertag{O-type} denotes all the other tokens~\cite{yang2020simple,ding2021few}.

\paragraph{Evaluation Scheme} To compare with SOTA models in Few-NERD leaderboard~\cite{ding2021few}, we adpot \emph{episode evaluation} as done by the authors. Here, a model is assessed by calculating the micro-F1 score over multiple number of test episodes. Each episode consists of a \emph{K-shot} support set and a \emph{K-shot} unlabeled query (test) set to make predictions. While Few-NERD is explicitly designed for episode evaluation, traditional NER datasets (e.g., OntoNotes, CoNLL'03, WNUT '17, GUM) have their distinctive tag-set distributions. Thus, sampling test episodes from the actual test data perturbs the true distribution that may not represent the actual performance. Consequently, \citet{yang2020simple} proposed to sample multiple support sets from the original development set and use them for prediction in the original test set. We also use this evaluation strategy for these traditional NER datasets.
\section{Method}

\ours utilizes contrastive learning to optimize distributional divergence between different token entity representations. Instead of focusing on label specific attributes, this contradistinction explicitly trains the model to distinguish between different categories of tokens. {Furthermore, modeling Gaussian Embedding instead of traditional point representation effectively lets \ours model the entity class distribution, which incites generalized representation of tokens.} Finally, it lets us carefully finetune our model even with a small number of samples without overfitting which is imperative for domain adaptation.

As demonstrated in Figure~\ref{fig:fig_architecture}, we first \textbf{train} our model in source domains. Next, we \textbf{finetune} model representations using few-sample support sets to adapt it to target domains. The training and finetuning of \ours is illustrated in Algorithm~\ref{algo:training}. Finally, we use an \textbf{instance level nearest neighbor classifier} for inference in test sets. 

\subsection{Model}
Figure~\ref{fig:fig_architecture} shows the key components of our model. To generate contextualized representation of sentence tokens, \ours incorporates a pretrained language model encoder \nertag{PLM}. For proper comparison against existing approaches, we use BERT~\cite{devlin2018bert} as our \nertag{PLM} encoder. Thus given a sequence of $n$ tokens $[x_1, x_2, \ldots, x_n]$, we take the final hidden layer output of the \nertag{PLM} as the intermediate representations $\bm{h}_i \in \mathbb{R}^{l^\prime}$.
\begin{equation}
    \label{eq:01}
    [\bm{h}_1, \bm{h}_2, \ldots, \bm{h}_n] = \nertag{PLM}([x_1, x_2, \ldots, x_n])
\end{equation}
These intermediate representations are then channeled through simple projection layer for generating the embedding. Unlike SimCLR \cite{chen2020simple} that uses projected point embedding for contrastive learning, we assume that token embeddings follow Gaussian distributions.
Specifically, we employ projection network $f_\mu$ and $f_\Sigma$ for producing Gaussian distribution parameters: 
\begin{equation}
    \label{eq:02}
    \bm{\mu}_i = f_\mu(\bm{h}_i),~~\bm{\Sigma}_i  = \elu\left(f_\Sigma(\bm{h}_i)\right) + (1 + \epsilon)
\end{equation}
where $\bm{\mu}_i \in \mathbb{R}^{l}, \bm{\Sigma}_i \in \mathbb{R}^{l \times l}$ represents mean and diagonal covariance (with nonzero elements only along the diagonal of the matrix) of the Gaussian Embedding respectively; 
$f_\mu$ and $f_\Sigma$ are implemented as ReLU followed by single layer networks;
$\elu$ for exponential linear unit; and $\epsilon \approx e^{-14}$ for numerical stability.

\subsection{Training in Source Domain} 
For calculating the contrastive loss, we consider the KL-divergence between all valid token pairs in the sampled batch. Two tokens $x_p$ and $x_q$ are considered as positive examples if they have the same label $y_p = y_q$. 
Given their Gaussian Embeddings $\mathcal{N}(\bm{\mu}_p,  \bm{\Sigma}_p)$ and $\mathcal{N}(\bm{\mu}_q,  \bm{\Sigma}_q)$, we can calculate their KL-divergence as following:
\begin{equation}
\label{eq:03}
\begin{split}
D_{\text{KL}}&[\mathcal{N}_q || \mathcal{N}_p] = D_{\text{KL}}[\mathcal{N}(\bm{\mu}_q,  \bm{\Sigma}_q) || \mathcal{N}(\bm{\mu}_p,  \bm{\Sigma}_p)] \\
& = \frac{1}{2}\biggl(\Tr( \bm{\Sigma}_p^{-1} \bm{\Sigma}_q) \\ & \quad \quad + (\bm{\mu}_p - \bm{\mu}_q)^T \bm{\Sigma}_p^{-1}(\bm{\mu}_p - \bm{\mu}_q) \\ & \quad \quad - l + \log\frac{| \bm{\Sigma}_p|}{| \bm{\Sigma}_q|}\biggr)
\end{split}
\end{equation}
Both directions of the KL-divergence are calculated since it is not symmetric.
\begin{equation}
    \label{eq:04}
    d(p,q) = \frac{1}{2}\left(D_{\text{KL}}[\mathcal{N}_q || \mathcal{N}_p] +  D_{\text{KL}}[\mathcal{N}_p || \mathcal{N}_q]\right)
\end{equation}
We first train our model in resource rich source domain having training data $\mathcal{X}_{\text{tr}}$. At each training step, we randomly sample a batch of sequences (without replacement) $\mathcal{X} \in \mathcal{X}_{\text{tr}}$ from the training set having batch size of $b$. For each $(x_i, y_i) \in \mathcal{X}$, we obtain its Gaussian Embedding $\mathcal{N}(\mu_i, \Sigma_i)$ by channeling the corresponding token sequence through the model (Algorithm \ref{algo:training}: Line 3-6). We find in-batch positive samples $\mathcal{X}_p$ for sample $p$ and subsequently calculate the Gaussian embedding loss of $x_p$ with respect to that of all other valid tokens in the batch:
 \begin{equation}
 \begin{split}
    \label{eqn:05}
     \mathcal{X}_p = \{(x_q, y_q) \in \mathcal{X} \mid y_p = y_q, p \neq q\}\\
\end{split}
 \end{equation}
 
 \begin{equation}
    \label{eqn:06}
     \ell(p) = -\log\frac{\sum\limits_{\substack{(x_q,y_q) \in \mathcal{X}_p}} \exp(-d(p,q)) / |\mathcal{X}_p|}{\sum\limits_{\substack{(x_q,y_q) \in \mathcal{X}, p \neq q}} \exp(-d(p,q))}
 \end{equation}
 In this way we can calculate the distributional divergence of all the token pairs in the batch (Algorithm \ref{algo:training}: Line 7-10 ).
 We do not scale the contrastive loss by any normalization factor as proposed by \citet{chen2020simple} since we did not find it to be beneficial for optimization.

 \subsection{Finetuning to Target Domain using Support Set}
 After training in source domains, we finetune our model using a small number of target domain support samples following a similar procedure as in the training stage. As we have only a few samples for finetuning, we take them in a single batch. When multiple few-shot samples (e.g., 5-shot) are available for the target classes, the model can effectively adapt to the new domain by optimizing KL-divergence of Gaussian Embeddings as in Eq. \ref{eq:04}. In contrast, for 1-shot case, it turns out challenging for models to adapt to the target class distribution. If the model has no prior knowledge about target classes (either from direct training or indirectly from source domain training where the target class entities are marked as \nertag{O-type}), a single example might not be sufficient to deduce the variance of the target class distribution. Thus, for 1-shot scenario, we optimize $d^\prime(p, q) = ||\bm{\mu}_p - \bm{\mu}_q||_2^2$, the squared euclidean distance between mean of the embedding distributions.
When the model has direct/indirect prior knowledge about the target classes involved, we still optimize the KL-divergence of the distributions similar to the 5-shot scenario. 

We demonstrate in Table \ref{tab:ft_obj} that optimizing with squared euclidean distance gives us slightly better performance in 1-shot scenario. Nevertheless, in all cases with 5-shot support set, optimizing the KL-divergence between the Gaussian Embeddings gives us the best result.

\paragraph{Early Stopping} Finetuning with a small support set runs the risk of overfitting and without access to a held out validation set due to data scarcity in the target domain, we cannot keep tabs on the saturation point where we need to stop finetuning. To alleviate this, we rely on the calculated contrastive loss and {use it as our early stopping criteria with a patience of 1. (Algorithm \ref{algo:training}: Line 16-17, 24 )} %
 
\begin{algorithm}[t!]
\caption{Training and Finetuning of \ours}
\begin{footnotesize}
\label{algo:training}
\begin{algorithmic}[1]
\REQUIRE Training data $\mathcal{X}_{tr}$, Support Data $\mathcal{X}_{sup}$, Train loss function $d_{tr}$, Finetune loss function $d_{ft}$ , $f_\mu, f_\Sigma$, \nertag{PLM}
\STATE \textcolor{blue}{// \emph{training in source domain}}
\FOR{sampled (w/o replacement) minibatch $\mathcal{X} \in \mathcal{X}_{tr}$ }
\FOR{\textbf{all} $i \equiv (x_i, y_i) \in \mathcal{X}$}
    \STATE $\bm{\mu}_i = f_{\mu}(\nertag{PLM}(x_i))$ \textcolor{blue}{\emph{//[Eq. \ref{eq:01}]}}
    \STATE $ \bm{\Sigma}_i = ELU(f_{\Sigma}(\nertag{PLM}(x_i))) + (1 + \epsilon)$ \textcolor{blue}{\emph{//[Eq. \ref{eq:02}]}}

\ENDFOR

\FOR{all $i \equiv (x_i, y_i) \in \mathcal{X}$}
    \STATE Calculate $\ell(i)$ as in Eq. \ref{eqn:05} and \ref{eqn:06}

\ENDFOR

\STATE $\mathcal{L}_{tr} = \frac{1}{|\mathcal{X}|} \sum\limits_{i \in \mathcal{X}} \ell(i)$
\STATE update $f_\mu, f_\Sigma,$ \nertag{PLM} by backpropagation to reduce $\mathcal{L}_{tr}$

\ENDFOR
\STATE \textcolor{blue}{// \emph{finetuning to target domain}}

\STATE $\mathcal{L}_{prev} = \infty$
\STATE $\mathcal{L}_{ft} = \mathcal{L}_{prev} - 1$ \textcolor{blue}{\emph{//Stable Initialization}}
\WHILE {$\mathcal{L}_{ft} < \mathcal{L}_{prev}$}
\STATE $\mathcal{L}_{prev} = \mathcal{L}_{ft}$
\FOR{\textbf{all} $i \equiv (x_i, y_i) \in \mathcal{X}_{sup}$}
    \STATE Calculate $\bm{\mu}_i$ and $ \bm{\Sigma}_i$ using Eq. \ref{eq:01}, \ref{eq:02} \textcolor{blue}{\emph{//Line 4,5}}

\ENDFOR
\FOR{all $i \equiv (x_i, y_i) \in \mathcal{X}_{sup}$}
\STATE Calculate $\ell(i)$ as in Eq. \ref{eqn:05} and \ref{eqn:06}
\ENDFOR
\STATE $\mathcal{L}_{ft} = \frac{1}{|\mathcal{X}_{sup}|} \sum\limits_{i \in \mathcal{X}_{sup}} \ell(i)$
\STATE update $f_\mu, f_\Sigma,$ \nertag{PLM} by backpropagation to reduce $\mathcal{L}_{ft}$
\ENDWHILE
\RETURN $\nertag{PLM}$ and discard $f_\mu$, $f_\Sigma$
\end{algorithmic}
\end{footnotesize}
\end{algorithm}

 \subsection{Instance Level Nearest Neighbor Inference} 
 \label{sec:instance_nn}
 
After training and finetuning the network with train and support data respectively, we extract the pretrained language model encoder \nertag{PLM} for inference. Similar to SimCLR~\cite{chen2020simple}, we found that representations before the projection layers actually contain more information than the final output representation which contributes to better performance, so $f_\mu$ and $f_\Sigma$ projection heads are not used for inference. We thus calculate the representations of the test data from \nertag{PLM} and find nearest neighbor support set representation for inference~\cite{wang2019simpleshot, yang2020simple}.

The \nertag{PLM} representations $\bm{h}_j^{\text{sup}}$ of each of the support token $(x_j^{\text{sup}},y_j^{\text{sup}}) \in \mathcal{X}_\text{sup}$ can be calculated as in Eq. \ref{eq:01}. Similarly for test data $\mathcal{X}_\text{test}$, we get the \nertag{PLM} representations $\bm{h}_i^\text{test}$ where $x_i^\text{test} \in \mathcal{X}_\text{test}$. Here we assign $x_i^\text{test}$ the same label as the support token that is nearest in the \nertag{PLM} representation space:
\begin{equation}
    \label{eq:07}
    y_i^\text{test} = \argmin\limits_{y_k^\text{sup}~\text{where}~ (x_k^\text{sup}, y_k^\text{sup}) \in \mathcal{X}_\text{sup}} ||\bm{h}_i^\text{test} - \bm{h}_k^\text{sup}||_2^2
\end{equation}

\begin{table}[H]
\centering
\small
\begin{tabular}{lrrrr}
    \toprule
     Dataset &  Domain & \# Class & \# Sent \\ \midrule
     OntoNotes & General & 18 & 76K \\
     I2B2'14 & Medical & 23 & 140K \\
     CoNLL'03 & News & 4 & 20K \\
     WNUT'17 & Social & 6 & 5K \\
     GUM & Mixed & 11 & 3.5K \\
     \textsc{Few}-NERD & Wikipedia & 66 & 188K \\
    \bottomrule
\end{tabular}
\caption{Summary Statistics of Datasets}
\label{tab:data}
\end{table}
\paragraph{Viterbi Decoding} Most previous works \cite{hou2020few, yang2020simple, ding2021few} noticed a performance improvement by using CRFs~\cite{lafferty2001conditional} which removes false predictions to improve performance. Thus we also employ Viterbi decoding in the inference stage with an abstract transition distribution as in StructShot~\cite{yang2020simple}. For the \textbf{transition probabilities}, the transition between three abstract tags \nertag{O}, \nertag{I}, and \nertag{I-other} is estimated by counting their occurrences in the training set. Then for the target domain tag-set, these transition probabilities are evenly distributed into corresponding target distributions. The \textbf{emission probabilities} are calculated from Nearest Neighbor Inference stage. Comparing domain transfer results (Table \ref{tab:domaintransfer}) against other tasks (Table \ref{tab:tagext},\ref{tab:fewnerd_intra},\ref{tab:fewnerd_inter}) we find that, interestingly, if there is no significant domain shift involved in the test data, contrastive learning allows \ours to automatically extract label dependencies, obviating the requirement of extra Viterbi decoding stage.

\section{Experiment Setups}
\label{sec:expts}
\paragraph{Datasets}
For evaluation, we use datasets across different domains: General (OntoNotes 5.0 \cite{weischedel2013ontonotes}), Medical (I2B2 \cite{stubbs2015annotating}), News (CoNLL'03 \cite{sang2003introduction}), Social (WNUT'17 \cite{derczynski2017results}). We also test on GUM~\cite{Zeldes2017} that represents wide variety of texts: interviews, news articles, instrumental texts, and travel guides. The miscellany of domains makes it a challenging dataset to work on.
\citet{ding2021few} argue that the distribution of these datasets may not be suitable for proper representation of Few-Shot capability. Thus, they proposed a new large scale dataset Few-NERD that contains 66 fine-grained entities across 8 coarse grained entities, significantly richer than previous datasets. A summary of these datasets is given in Table \ref{tab:data}.

\paragraph{Baselines}

\begin{table*}[t]
\centering
\scalebox{.80}{%
\begin{tabular}{lcccccccccc}
    \toprule
    \multirow{2}{*}{\textbf{Model}} & \multicolumn{3}{c}{\textbf{1-shot}} &  & \multicolumn{3}{c}{\textbf{5-shot}} & \\
    \cmidrule(r){2-5} \cmidrule(r){6-9} 
    & \textbf{Group A} & \textbf{Group B} & \textbf{Group C} & \textbf{Avg.} & \textbf{Group A} & \textbf{Group B} & \textbf{Group C} & \textbf{Avg.} \\
    
    Proto & 19.3 $\pm$ 3.9 & 22.7 $\pm$ 8.9 & 18.9 $\pm$ 7.9 & 20.3 & 30.5 $\pm$ 3.5 & 38.7 $\pm$ 5.6 & 41.1 $\pm$ 3.3 & 36.7 \\
    NNShot & 28.5 $\pm$ 9.2 & 27.3 $\pm$ 12.3 & 21.4 $\pm$ 9.7 & 25.7 & 44.0 $\pm$ 2.1 & 51.6 $\pm$ 5.9 & 47.6 $\pm$ 2.8 & 47.7  \\
    StructShot & 30.5 $\pm$ 12.3 & 28.8 $\pm$ 11.2 & 20.8 $\pm$ 9.9 & 26.7 & 47.5 $\pm$ 4.0 & 53.0 $\pm$ 7.9 & 48.7 $\pm$ 2.7 & 49.8 \\
    \textbf{CONTaiNER} & \textbf{32.2 $\pm$ 5.3} & \textbf{30.9 $\pm$ 11.6} & \textbf{32.9 $\pm$ 12.7} & \textbf{32.0} &  \textbf{51.2 $\pm$ 5.9} & \textbf{55.9 $\pm$ 6.2} & \textbf{61.5 $\pm$ 2.7} & \textbf{56.2}\\
    \textbf{\quad + Viterbi} & \textbf{32.4 $\pm$ 5.1} & \textbf{30.9 $\pm$ 11.6} & \textbf{33.0 $\pm$ 12.8} & \textbf{32.1} & \textbf{51.2 $\pm$ 6.0} & \textbf{56.0 $\pm$ 6.2} & \textbf{61.5 $\pm$ 2.7} & \textbf{56.2}\\
    
    \bottomrule
\end{tabular}}
\caption{\label{tab:tagext}F1 scores in Tag Set Extension on OntoNotes. Group A, B, C are three disjoint sets of entity types. Results vary slightly compared to \citet{yang2020simple} since they used different support set samples (publicly unavailable) than ours.
}
\vspace{-3mm}
\end{table*}
\begin{table*}[t]
\centering
\scalebox{.70}{%
\begin{tabular}{lcccccccccc}
    \toprule
    \multirow{2}{*}{\textbf{Model}} & \multicolumn{5}{c}{\textbf{1-shot}} &  \multicolumn{5}{c}{\textbf{5-shot}} \\
    \cmidrule(r){2-6} \cmidrule(r){7-11} 
    & \textbf{I2B2} & \textbf{CoNLL} & \textbf{WNUT} & \textbf{GUM} & \textbf{Avg.} & \textbf{I2B2} & \textbf{CoNLL} & \textbf{WNUT} & \textbf{GUM} & \textbf{Avg.} \\

    Proto & 13.4 $\pm$ 3.0 & 49.9 $\pm$ 8.6 & 17.4 $\pm$ 4.9 & 17.8 $\pm$ 3.5 & 24.6 & 17.9 $\pm$ 1.8 & 61.3 $\pm$ 9.1 & 22.8 $\pm$ 4.5 & 19.5 $\pm$ 3.4 & 30.4\\

    NNShot & 15.3 $\pm$ 1.6 & 61.2 $\pm$ 10.4 & 22.7 $\pm$ 7.4 & 10.5 $\pm$ 2.9 & 27.4 & 22.0 $\pm$ 1.5 & 74.1 $\pm$ 2.3 & 27.3 $\pm$ 5.4 & 15.9 $\pm$ 1.8 & 34.8\\

    StructShot & 21.4 $\pm$ 3.8 & \textbf{62.4 $\pm$ 10.5} & 24.2 $\pm$ 8.0 & 7.8 $\pm$ 2.1 & 29.0  & 30.3 $\pm$ 2.1 & 74.8 $\pm$ 2.4 & 30.4 $\pm$ 6.5 & 13.3 $\pm$ 1.3 & 37.2 \\

    \textbf{CONTaiNER} & {16.4 $\pm$ 1.7} & {57.8 $\pm$ 10.7} &{24.2 $\pm$ 2.9} & {17.9 $\pm$ 1.8} & {29.1} &  {24.1 $\pm$ 1.9} & {72.8 $\pm$ 2.0} & {27.7 $\pm$ 2.2}  & {24.4 $\pm$ 2.2} & {37.3}\\

    \textbf{\quad + Viterbi} & \textbf{21.5 $\pm$ 1.7} & {61.2 $\pm$ 10.7} & \textbf{27.5 $\pm$ 1.9} & \textbf{18.5 $\pm$ 4.9} & \textbf{32.2} & \textbf{36.7 $\pm$ 2.1} & \textbf{75.8 $\pm$ 2.7} & \textbf{32.5 $\pm$ 3.8} & \textbf{25.2 $\pm$ 2.7} & \textbf{42.6}\\
    \bottomrule
\end{tabular}}
\caption{\label{tab:domaintransfer}F1 scores in Domain Extension with OntoNotes as the source domain. Results vary slightly compared to \citet{yang2020simple} since they used different support set samples (publicly unavailable) than ours. %
}
\vspace{-4mm}
\end{table*}

We compare the performance of \ours with state-of-the-art Few-Shot NER models on different datasets across several settings. We first measure the model performance in traditional NER datasets in tag-set extension and domain transfer tasks as proposed in~\citet{yang2020simple}. We then evaluate our model in Few-NERD~\cite{ding2021few} dataset that is explicitly designed for Few-Shot NER, and compare it against the Few-NERD leaderboard baselines. Similar to~\citet{ding2021few}, we take Prototypical Network based ProtoBERT~\cite{snell2017prototypical, fritzler2019few, hou2020few}, nearest neighbor based metric method NNShot {that leverages the locality of in-class samples in embedding space}, and additional Viterbi decoding based Structshot \cite{yang2020simple} as the main SOTA baselines.

\subsection{Tag-set Extension Setting}

A common use-case of Few-Shot NER is that new entity types may appear in the same existing text domain. Thus \citep{yang2020simple} proposed to experiment tag-set extension capability using OntoNotes \cite{weischedel2013ontonotes} dataset. The eighteen existing entity classes are split in three groups: A, B, and C, each having six classes. Models are tested in each of these groups having few sample support set while being trained in the remaining two groups. During training, all test group entities are replaced with \nertag{O}-tag. Since the source and destination domains are the same, the training phase will induce some indirect information about unseen target entities. So, during finetuning of \ours, we optimize the KL-divergence between ouptut embeddings as in Eq. \ref{eq:04}.

We use the same entity class splits as used by \citet{yang2020simple} and used \nertag{bert-base-cased} as the backbone encoder for all models. Since they could not share the sampled support set for licensing reasons, we sampled five sets of support samples for each group and averaged the results, as done by the authors. We show these results in Table \ref{tab:tagext}. We see that in different entity groups, \ours outperforms present SOTAs by upto 12.75 absolute F1 points, a substantial improvement in performance.

\subsection{Domain Transfer Setting}
In this experiment a model trained on a source domain is deployed to a previously unseen novel text domain. Here we take OntoNotes (General) as our source text domain, and evaluate the Few-Shot performance in I2B2 (Medical), CoNLL (News), WNUT (Social) domains as in \cite{yang2020simple}. We also evaluate the performance in GUM \cite{Zeldes2017} dataset due to its particularly challenging nature. %
We show these results in Table \ref{tab:domaintransfer}. While all the other domains have almost no intersection with OntoNotes, target entities in CoNLL are fully contained within OntoNotes entities, that makes it comparable to supervised learning. %

\subsection{Few-NERD Setting}

For few-shot setting, \citet{ding2021few} proposed two different settings: \textbf{Few-NERD (INTRA)} and \textbf{Few-NERD (INTER)}. In Few-NERD (INTRA) train, dev, and test sets are divided according to coarse-grained types. As a result, fine-grained entity types belonging to \nertag{People, Art, Product, MISC} coarse grained types are put in the train set, \nertag{Event, Building} coarse grained types in dev set, and \nertag{ORG, LOC} in test set. So, there is no overlap between train, dev, test set classes in terms of coarse grained types. On the other hand, in Few-NERD (INTER) coarse grained types are shared, although all the fine grained types are mutually disjoint. Because of the restrictions of sharing coarse-grained types, Few-NERD (INTRA) is more challenging.
\begin{table}[t!]
\centering
\scalebox{.60}{%
\begin{tabular}{lcccccr}
    \toprule
    \multirow{2}{*}{\textbf{Model}} & \multicolumn{2}{c}{\textbf{5-way}} & \multicolumn{2}{c}{\textbf{10-way}} & \multirow{2}{*}{\textbf{Avg.}}\\
    \cmidrule(r){2-3} \cmidrule(r){4-5}
    & \textbf{1$\sim$2 shot} & \textbf{5$\sim$10 shot} & \textbf{1$\sim$2 shot} & \textbf{5$\sim$10 shot} \\
    \cmidrule(r){1-1} \cmidrule(r){2-3} \cmidrule(r){4-5} \cmidrule(r){6-6}
    StructShot & 35.92 & 38.83 & 25.38 & 26.39 & 31.63 \\
    ProtoBERT & 23.45 & 41.93 & 19.76 & 34.61 & 29.94 \\
    NNShot & 31.01 & 35.74 & 21.88 & 27.67 & 29.08 \\
    \textbf{CONTaiNER} & \textbf{40.43} & \textbf{53.70} & \textbf{33.84} & \textbf{47.49} & \textbf{43.87} \\
    {\textbf{\quad + Viterbi}} & \textbf{40.40} & \textbf{53.71} & \textbf{33.82} & \textbf{47.51} & \textbf{43.86}\\

    \bottomrule
\end{tabular}}
\caption{F1 scores in FEW-NERD (INTRA). 
}
\label{tab:fewnerd_intra}
\vspace{-4mm}
\end{table}
\begin{table}[t!]
\centering
\scalebox{.60}{%
\begin{tabular}{lcccccr}
    \toprule
    \multirow{2}{*}{\textbf{Model}} & \multicolumn{2}{c}{\textbf{5-way}} & \multicolumn{2}{c}{\textbf{10-way}} & \multirow{2}{*}{\textbf{Avg.}}\\
    \cmidrule(r){2-3} \cmidrule(r){4-5}
    & \textbf{1$\sim$2 shot} & \textbf{5$\sim$10 shot} & \textbf{1$\sim$2 shot} & \textbf{5$\sim$10 shot} \\
    \cmidrule(r){1-1} \cmidrule(r){2-3} \cmidrule(r){4-5} \cmidrule(r){6-6}
    StructShot & \textbf{57.33} & 57.16 & \textbf{49.46} & 49.39 & 53.34 \\
    ProtoBERT & 44.44 & 58.80 & 39.09 & 53.97 & 49.08 \\
    NNShot & 54.29 & 50.56 & 46.98 & 50.00 & 50.46 \\
    \textbf{CONTaiNER} & 55.95 & \textbf{61.83} & 48.35 & \textbf{57.12} & \textbf{55.81} \\
    {\textbf{\quad + Viterbi}} & {56.1} & \textbf{61.90} & {48.36} & \textbf{57.13} & \textbf{55.87}\\
    
    %

    \bottomrule
\end{tabular}}
\caption{F1 scores in FEW-NERD (INTER). 
}
\label{tab:fewnerd_inter}
\vspace{-5mm}
\end{table}
\label{sec: experiment}
Since, few-shot performance of any model relies on the sampled support set, the authors also released train, dev, test split for both \textbf{Few-NERD (INTRA) } and \textbf{Few-NERD (INTER)}. We evaluate our model performance using these provided dataset splits and compare the performance in Few-NERD leaderboard.
All models use \nertag{bert-base-uncased} as the backbone encoder. As shown in Table \ref{tab:fewnerd_intra} and Table \ref{tab:fewnerd_inter}, \ours establishes new benchmark results in the leaderboard in both of these tests.

\section{Results and Analysis}
We prudently analyze different components of our model and justify the design choices made in the scheming of \ours. We also examine the results discussed in Section \ref{sec:expts} that gives  some intuitions about few-shot NER in general.

\subsection{Overall Results}
Table \ref{tab:tagext}-\ref{tab:fewnerd_inter} demonstrates that overall, in every scenario \ours convincingly outperforms all other baseline approaches. This improvement is particularly noticeable in challenging scenarios, where all other baseline approaches perform poorly. For example, FEW-NERD (intra) (Table \ref{tab:fewnerd_intra}) is a challenging scenario where the coarse grained entity types corresponding to train and test sets do not overlap. As a result, other baseline approaches face a substantial performance hit, whereas \ours still performs well. In tag-set extension (Table \ref{tab:tagext}), we see a similar performance trend - \ours performs consistently well across the board. Likewise, in domain transfer to a very challenging unseen text domain like GUM \cite{Zeldes2017}, baseline models performs miserably; yet \ours manages to perform consistently outperforming SOTA models by a significant margin. Analyzing these results more closely, we notice that while \ours surpasses other baselines in almost every tests, more prominently in 5-shot cases. Evidently, \ours is able to make better use of multiple few-shot samples thanks to distribution modeling via contrastive Gaussian Embedding optimization. In this context, note that StructShot actually got marginally higher F1-score in 1-shot CoNLL domain adaptation and 1$\sim$2 shot FEW-NERD (INTER) cases. In CoNLL, the target classes are subsets of training classes, so supervised learning based feature extractors are expected to get an advantage in prediction. On the other hand, \citet{ding2021few} carefully tuned the hyperparameters for baselines like StructShot for best performance. We could also improve performance in a similar manner, however for uniformity of model across different few-shot settings, we use the same model architecture in every test. Nevertheless, \ours shows comparable performance even in these cases while significantly outperforming in every other test.

\subsection{Training Objective}
\label{sec: train_obj}

Traditional contrastive learners usually optimize cosine similarity of point embeddings \cite{chen2020simple}. While this has proven to work well in image data, in more challenging NLU tasks like Few-Shot NER, it gives subpar performance. {We compare the performance of point embeddings with euclidean distance and cosine similarity to that of \ours using Gaussian Embedding and KL-divergence in OntoNotes tag-set extension. We report these performance in Table \ref{tab:tagext_obj} in Appendix. Basically, Gaussian Embedding leads to learning generalized representation during training, which is more suitable for finetuning to few sample target domain. In Appendix \ref{tsne_vis}, we examine this aspect by comparing the t-SNE representations from point embedding and Gaussian Embedding.} %

\subsection{Effect of Model Fine-tuning}
Being a contrastive learner, \ours can take advantage of extremely small support set to refine its representations through fine-tuning. To closely examine the effects of fine-tuning, we conduct a case study with OntoNotes tag-extension task using \nertag{PERSON, DATE, MONEY, LOC, FAC, PRODUCT} target entities. 

\begin{table}[h!]
\centering

\scalebox{.90}{%
\begin{tabular}{ccc}
    \toprule
    & W/O Finetuning & W/ Finetuning \\ 
    \midrule
   
    1-shot & 31.76 & 32.90 \\
    5-shot & 56.99 & 61.48 \\
    \bottomrule
\end{tabular}}
\caption{Comparison of F1-Scores with and without support set finetuning of \ours}
\label{tab:ft_effect}
\end{table}

As shown in Table \ref{tab:ft_effect}, we see that finetuning indeed improves few-shot performance. Besides, the effect of finetuning is even more marked in 5-shot prediction indicating that \ours finetuning process can make the best use of few-samples available in target domain.

\subsection{Modeling Label Dependencies}
Analyzing the results, we observe that domain transfer (Table \ref{tab:domaintransfer}) sees some good gains in performance from using Viterbi decoding. In contrast, tag-set extension (Table \ref{tab:tagext}) and FEW-NERD (Table \ref{tab:fewnerd_intra},\ref{tab:fewnerd_inter}) gets almost no improvement from using Viterbi decoding. This indicates an interesting property of \ours. During domain transfer the text domains have no overlap in train and test set. So, an extra Viterbi decoding actually provides additional information regarding the label dependencies, giving us some nice improvement. Otherwise, the train and target domain have substantial overlap in both tagset extension and FEW-NERD. Thus the model can indirectly learn the label dependencies through in-batch contrastive learning. Consequently, unless there is a marked shift in the target text domain, we can achieve the best performance even without employing additional Viterbi decoding.

\section{Related Works}
\paragraph{Meta Learning} %
The idea of Few-shot learning was popularized in computer vision through Matching Networks \cite{vinyals2016matching}. Subsequently, Prototypical Network \cite{snell2017prototypical} was proposed where class prototypical representations were learned. Test samples are given labels according to the nearest prototype. Later this technique was proven successful in other domains as well. \citet{wang2019simpleshot}, on the other hand found simple feature transformations to be quite effective in few shot image recognition  These metric learning based approaches have also been deployed in different NLP tasks \cite{geng2019induction, bao2019few, han2018fewrel, fritzler2019few}.

\paragraph{Contrastive Learning} 
Early progress was made by contrasting positive against negative samples \cite{hadsell2006dimensionality,dosovitskiy2014discriminative, wu2018unsupervised}. \citet{chen2020simple} proposed SimCLR by refining the idea of contrastive learning with the help of modern image augmentation techniques to learn robust sets of features. \citet{khosla2020supervised} leveraged this to boost supervised learning performance as well. In-batch negative sampling has also been explored for learning representation \cite{doersch2017multi,ye2019unsupervised}. Storing instance class representation vectors is another popular direction~\cite{wu2018unsupervised,zhuang2019local,misra2020self}.

\paragraph{Gaussian Embedding}
\citet{vilnis2014word} first explored the idea of  learning word embeddings as Gaussian Distributions. Although the authors used RANK-SVM based learning objective instead of modern deep contextual modeling, they found that embedding  densities in a Gaussian space enables natural represenation of uncertainty through variances. Later, \citet{bojchevski2017deep} leveraged Gaussian Embedding in Graph representation. Besides state-of-the-art performance, they found Gaussian Embedding to be surprisingly effective in \textbf{inductive} learning, generalizing to unseen nodes with few training data. Moreover, KL-divergence between Gaussian Embeddings allows explicit consideration of asymmetric distance which better represents inclusion, similarity or entailment \cite{qian2021conceptualized} and preserve the hierarchical structures among words \cite{athiwaratkun2018hierarchical}.

\paragraph{Few-Shot NER}

Established few-shot learning approaches have also been applied in Named Entity Recognition. \citet{fritzler2019few} leveraged prototypical network~\cite{snell2017prototypical} for few shot NER. Inspired by the potency of simple feature extractors and nearest neighbor inference \cite{wang2019simpleshot,wiseman2019label} in few-Shot learning, \citet{yang2020simple} used supervised learner based feature extractors for Few-Shot NER. Pairing it with abstract transition tag Viterbi decoding, they achieved current SOTA result in Few-Shot NER tasks. \citet{huang2020few} proposed noisy supervised pre-training for Few-Shot NER. However, this method requires access to a large scale noisy NER dataset such as WiNER \cite{ghaddar2017winer} for the supervised pretraining. Acknowledging the shortcomings and evaluation scheme disparity in Few-Shot NER, \citet{ding2021few} proposed a large scale dataset specifically designed for this task. \citet{wang2021meta} explored model distillation for Few-Shot NER. However, this requires access to a large unlabelled dataset for good performance. Very recently, prompt based techniques have also surfaced in this domain \cite{cui2021template}. However, the performance of these methods rely heavily on the chosen prompt. As denoted by the author, the performance delta can be massive (upto 19\% absolute F1 points) depending on the prompt. Thus, in the absence of a large validation set, their applicability becomes limited in true few-shot learning \cite{perez2021true}.
\section{Conclusion}
We propose a contrastive learning based framework \ours that models Gaussian embedding and optimizes inter token distribution distance. This generalized objective helps us model a class agnostic feature extractor that avoids the pitfalls of prior Few-Shot NER methods. \ours can also take advantage of few-sample support data to adapt to new target domains. Extensive evaluations in multiple traditional and recent few-shot NER datasets reveal that, \ours consistently outperforms prior SOTAs, even in challenging scenarios. While we investigate the efficacy of distribution optimization based contrastive learning in Few-Shot NER, it will be of particular interest to investigate its potency in other domains as well. 
\section*{Acknowledgement}
We thank the ACL Rolling Review reviewers for their helpful feedback. We also want to thank Nan Zhang, Ranran Haoran Zhang, and Chandan Akiti for their insightful comments on the paper.
\section*{Ethics Statement}
With \ours, we have achieved state-of-the-art Few-Shot NER performance leveraging Gaussian Embedding based contrastive learning. However, the overall performance is still quite low compared to supervised NER that takes advantage of the full training dataset. Consequently, it is still not ready for deployment in high-stake domains (e.g. Medical Domain, I2B2 dataset), leaving a lot of room for improvement in future research.
\bibliographystyle{acl_natbib}
\bibliography{bibil.bib}
\clearpage
\appendix
\section{Implementation Details}
For all of our experiments in \ours. we chose the same hyperparameters as in \citet{yang2020simple}. Across all our tests, we kept Gaussian Embedding dimension fixed to $l = 128$.  In order to guarantee proper comparison against prior competitive approaches, we use the same backbone encoder for all methods in same tests, i.e. \nertag{bert-base-cased} was used for all methods in Tag-Set Extension and Domain Transfer tasks while \nertag{bert-base-uncased} was used for Few-NERD following the respective evaluation strategies. Finally, to observe the effect of Viterbi decoding on \ours output, we set the re-normalizing temperature $\tau$ from [0.01, 0.05, 0.1]. 

Using an RTX A6000, we trained the network on OntoNotes dataset for 30 minutes. The finetuning stage requires less than a minute due to the small number of samples.

\section{Fine-tuning Objective}
During finetuning, if a model does not have any prior knowledge about the target classes, directly or indirectly, a 1-shot example may not give sufficient information about the target class distribution (i.e. the variance of the distribution). Consequently during finetuning, for 1-shot adaptation to new classes, optimizing euclidean distance of the mean embedding gives better performance. Nevertheless, for 5-shot cases, KL-divergence of the Gaussian Embedding always gives better performance indicating that it takes better advantage of multiple samples. We show this behavior in the best result of domain transfer task with WNUT in Table \ref{tab:ft_obj}. Since this domain transfer task gives no prior information about target embeddings during training, optimizing KL-divergence in 1-shot fineutuning actually hurts performance a bit compared to  euclidean finetuning. However, in 5-shot, KL-finetuning again gives superior performance as it can now adapt better to the novel target class distributions. 
\begin{table}[h!]
\centering

\scalebox{.90}{%
\begin{tabular}{ccc}
    \toprule
    
    & KL-Gaussian & Euclidean-mean \\ 
    \midrule
   
    1-shot & 18.78 & 27.48 \\
    5-shot & 32.50 & 31.12 \\
    
    \bottomrule
\end{tabular}}
\caption{F1 scores comparison in Domain Transfer Task with WNUT with different \emph{finetune} objectives. While optimizing the KL-divergence of the Gaussian Embedding gives superior result in 5-shot, optimizing Euclidean distance of the mean embeddings actually achieve better result in 1-shot. Note that in both cases the model is \emph{trained} on out-of-domain data using \textbf{KL-Gaussian}.}
\label{tab:ft_obj}
\end{table}

\begin{figure*}[h]
\includegraphics[width=\textwidth,height=0.45\textheight]{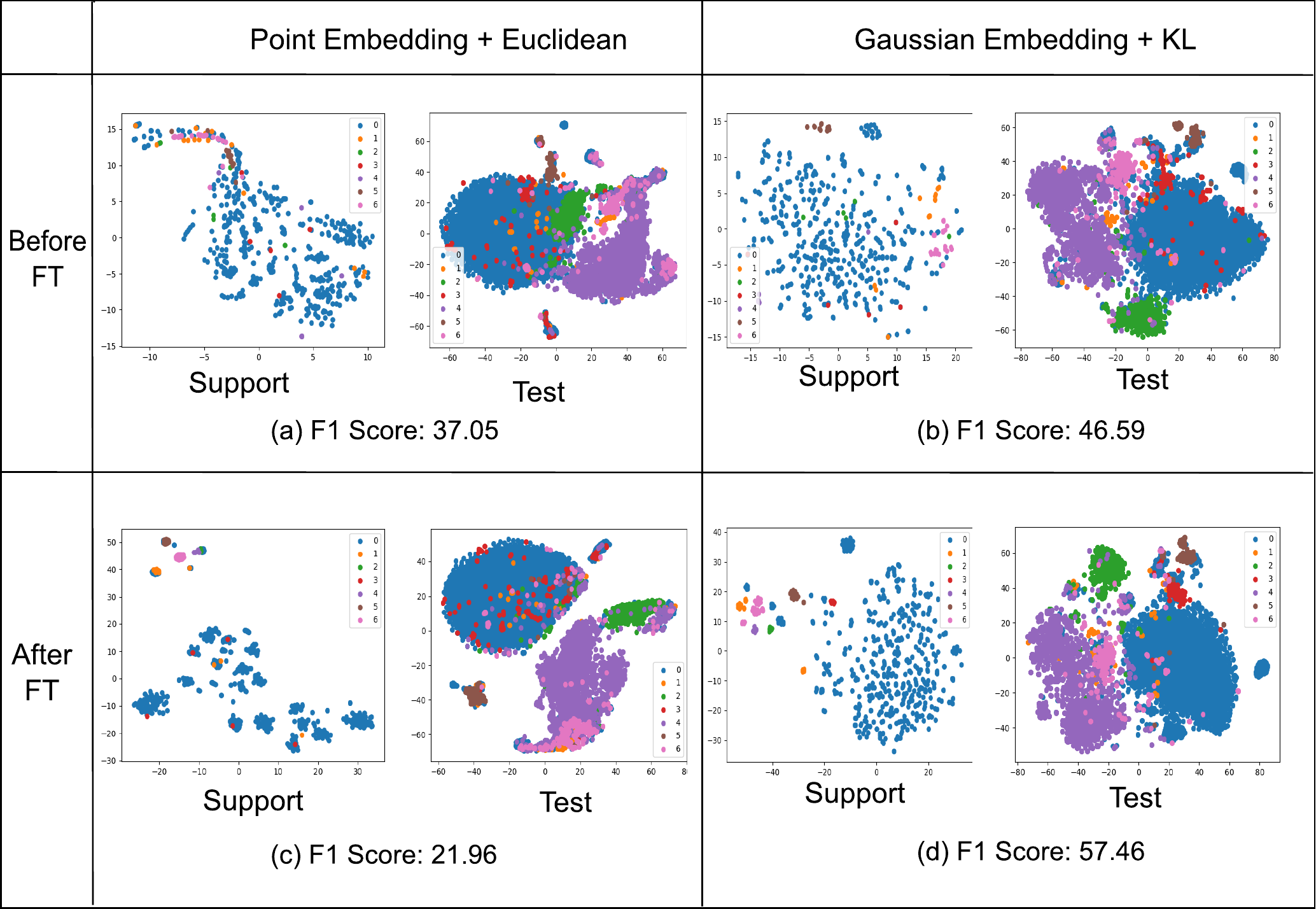}

\caption{t-SNE visualization of support set and test set representations in a sample few-shot task in OntoNotes tag extension. We show both support and test set representation here before and after finetuning. \textbf{Prior to finetuning, (a)} contrastive learner with point embedding and Euclidean distance objective gives intermixed class representations; \textbf{(b)} Gaussian Embedding with KL-divergenece generates clusters for different unseen classes. \textbf{After finetuning, (c)} point embedding overfits the support examples which further intermingles different class examples; \textbf{(d)} Gaussian Embedding with KL-divergence cleans up the clusters offering better separation between different classes, which results in higher F1-score.}

\label{fig:tsne}
\end{figure*}
{
\section{t-SNE Visualization: Point Embedding vs. Gaussian Embedding} \label{tsne_vis}
Figure \ref{fig:tsne} offers a deep dive into how Gaussian Embedding improves generalization and takes better advantage of few shot support set for target domain adaptation. Here we compare the t-SNE visualization of support set and test set of a sample few-shot scenario in OntoNotes tag set extension task. In Figure \ref{fig:tsne} (a) we can see that point embedding paired with Euclidean distance metric has suboptimal clustering pattern in both support and test sets. In fact, the support examples in different classes are intermixed implying poor generalization. When the point embedding model is finetuned with the support set (Figure \ref{fig:tsne} (c)), Euclidean distance aggressively optimizes them and tries to force the same class support examples to collapse into essentially a single point representation. In other words, the model quickly overfits the small support data which in fact hurts model performance.}
{
In comparsion, Gaussian Embedding offers a better t-SNE representation prior to and after finetuning. Figure \ref{fig:tsne} (b) shows the representation of support and test sets prior to finetuning with Gaussian Embedding paired with KL-divergence. In both support and test sets, we observe different class samples mostly clustered together. This indicates that even before finetuning it shows good generalization to unseen classes. While finetuning, the KL-divergence optimization objective maintains the class distribution letting the model generate separate support clusters (Figure \ref{fig:tsne}(d)). After finetuning, the clusters get cleaner offering even better separation between different class clusters, which is also reflected in the performance uplift of the model.}

\begin{table*}[!t]
\centering
\scalebox{.84}{%
\begin{tabular}{lcccccccccc}
    \toprule
    \multirow{2}{*}{\textbf{Model}} & \multicolumn{3}{c}{\textbf{1-shot}} &  & \multicolumn{3}{c}{\textbf{5-shot}} & \\
    \cmidrule(r){2-5} \cmidrule(r){6-9} 
    & \textbf{Group A} & \textbf{Group B} & \textbf{Group C} & \textbf{Avg.} & \textbf{Group A} & \textbf{Group B} & \textbf{Group C} & \textbf{Avg.} \\
    
    Point Embedding + Cosine & 7.73 & 11.27 & 15.57 & 11.52 & 17.33 & 30.08 & 22.51 & 23.31\\
    Point Embedding + Euclidean & 14.96 & 13.67 & 11.12 & 13.25 & 25.35 & 41.56 & 43.11 & 36.67  \\

    Gaussian Embedding + KL-div. & \textbf{32.2} & \textbf{30.9} & \textbf{32.9} & \textbf{32.0} &  \textbf{51.2} & \textbf{55.9} & \textbf{61.5} & \textbf{56.2}\\

    \bottomrule
\end{tabular}}
\caption{ OntoNotes Tag Set extension mean-F1 score comparison between Point Embedding (with Euclidean distance and cosine similarity) and Gaussian Embedding (KL-divergence). 
}
\label{tab:tagext_obj}
\vspace{-2mm}
\end{table*}

\section{Comparison of Different Training Objectives}
Table \ref{tab:tagext_obj} compares the performance of Gaussian Embedding (KL-divergence) with that of point embedding (Euclidean distance of cosine similarity) in OntoNotes tag extension task. Since Gaussian Embedding utilizes $l$ dimensional mean and $l$ dimensional diagonal covariance matrix, for a fair comaparison we show the results for $2l$ dimensional point embedding. As discussed in Section \ref{sec: train_obj}, Gaussian Embedding with KL-divergence objective largely outperforms point embedding irrespective of distance metric used.

\section{Embedding Quality: Before vs. After Projection}
\begin{table}[h!]
\centering

\scalebox{.85}{%
\begin{tabular}{ccc}
    \toprule
    & Before Projection & After Projection \\ 
    \midrule
   
    1-shot & \textbf{32.17} & 29.21 \\
    5-shot & \textbf{51.19} & 49.78 \\
    \bottomrule
\end{tabular}}
\caption{Comparison of F1-Scores on OntoNotes Group A before and after the projection layer of \ours}
\label{tab:projection_compare}
\vspace{-3mm}
\end{table}
As explained in Section \ref{sec:instance_nn}, the representation before the projection layer contains more information than that of after. In Table \ref{tab:projection_compare}, we compare the performance of representations before and after the Gaussian projection layer. From the results it is evident that, representation before the projection indeed achieves higher performance, which also supports the findings of \cite{chen2020simple}. This is because the representation after the projection head is directly adjacent to the contrastive objective, which causes information  loss in this layer. Consequently, the representation before projection achieves better performance.

\section{NER Prediction Examples}

Table \ref{tab:case_study} demonstrates some predictions with \ours and StructShot using \nertag{PERSON, DATE, MONEY, LOC, FAC, PRODUCT} as target few-shot entities while being trained on all other entity types in OntoNotes dataset. {A quick look at these qualitative examples reveal that StructShot often fails to distinguish between non-entity and entity tokens. Moreover, it also misclassifies non-entity tokens as one of the target classes. \ours on the other hand has lower misclassifications and better entity detection indicating its stability and higher performance.}

\onecolumn

\footnotesize
\begin{longtable}{p{5cm} p{5cm} p{5cm}}
\toprule
\textbf{Gold} & \textbf{\textsc{CONTaiNER}} & \textbf{StructShot} \\
\midrule 
BMEC general director Dr. \tokentag{Johnsee Lee}{PER} says that the ITRI 's \tokentag{four-year}{DATE} R\&D program in biochip applications and technology is now in its \tokentag{second year}{DATE}. & BMEC general director Dr. \tokentag{Johnsee Lee}{PER} says that the ITRI 's \tokentag{four-year}{DATE} R\&D program in biochip applications and technology is now in its \tokentag{second year}{DATE}. & BMEC general director Dr. \tokentag{Johnsee Lee}{PER} says that the ITRI 's \tokentag{four-year}{DATE} R\&D program in biochip applications and technology is now in its second year.\\
\midrule
DR. Chip Bio-technology was set up in \tokentag{September 1998}{DATE}. & DR. Chip Bio-technology was set up in \tokentag{September 1998}{DATE}. & \tokentag{DR. Chip Bio-technology}{PRODUCT} was set up in {September 1998}.\\
\midrule
\tokentag{Wang Shin - hwan}{PER} notes that traditional bacterial and viral cultures take seven to ten days to prepare , and even with the newer molecular biology testing techniques it takes \tokentag{three days}{DATE} to get a result . & \tokentag{Wang Shin}{PER}  - hwan notes that traditional bacterial and viral cultures take \tokentag{seven to ten days}{DATE} to prepare , and even with the newer molecular biology testing techniques it takes \tokentag{three days}{DATE} to get a result . & Wang Shin - hwan notes that traditional bacterial and viral cultures take \tokentag{seven to ten days}{DATE} to prepare , and even with the newer molecular biology testing techniques it takes \tokentag{three days}{DATE} to get a result .\\
\midrule
Research program director \tokentag{Pan Chao - chi}{PER} states that at present they are actively developing a " fever chip " with a wide range of applications . & Research program director \tokentag{Pan}{PER} Chao - chi states that at \tokentag{present}{DATE} they are actively developing a " fever chip " with a wide range of applications . & Research program director \tokentag{Pan}{PER} Chao - chi states that at present they are actively developing a " fever chip " with a wide range of applications .\\
\midrule
Pan explains that in clinical practice , the causes of fever are difficult to quickly diagnose . & \tokentag{Pan}{PER} explains that in clinical practice , the causes of fever are difficult to quickly diagnose . & Pan explains that in clinical practice , the causes of fever are difficult to quickly diagnose . \\
\midrule
\tokentag{Jerry Huang}{PER}, executive vice president of U - Vision Biotech , reveals that U - Vision , which was set up in \tokentag{September 1999}{DATE}, has signed a contract with the US company Zen - Bio to jointly develop human adipocyte cDNA microarray chips . & \tokentag{Jerry Huang}{PER}, executive vice president of U - Vision Biotech , reveals that U - Vision , which was set up in \tokentag{September 1999}{DATE}, has signed a contract with the US company Zen - Bio to jointly develop human adipocyte cDNA microarray chips . & Jerry Huang , executive vice president of U - Vision Biotech , reveals that U - Vision , which was set up in September 1999 , has signed a contract with the US company Zen - Bio to jointly develop human adipocyte cDNA microarray chips. \\
\midrule
\tokentag{Huang}{PER} states that research in \tokentag{recent years}{DATE}has revealed that adipocytes -LR fat cells -RR are active regulators of the energy balance in the body , and play an important role in disorders such as obesity , diabetes , osteoporosis and cardiovascular disease . & \tokentag{Huang}{PER} states that research in \tokentag{recent years}{DATE}has revealed that adipocytes -LR fat cells -RR are active regulators of the energy balance in the body , and play an important role in disorders such as obesity , diabetes , osteoporosis and cardiovascular disease . & Huang states that research in recent years has revealed that adipocytes -LR fat cells -RR are active regulators of the energy balance in the body , and play an important role in disorders such as obesity , diabetes , osteoporosis and cardiovascular disease .\\
\midrule
Maybe a \tokentag{30 year old}{DATE} man \& a \tokentag{15 year old}{DATE}boy doesn't qualify . & Maybe a 30 year old man \& a 15 year old boy doesn't qualify . & Maybe a \tokentag{30 year old}{DATE} man \& a \tokentag{15 year old}{DATE}boy doesn't qualify .\\
\midrule
After \tokentag{Tom DeLay}{PER} was zapped, \tokentag{Charles Colson}{PER} became \tokentag{DeLay}{PER}'s personal guru. & After \tokentag{Tom DeLay}{PER} was zapped, \tokentag{Charles Colson}{PER} became {DeLay}'s personal guru. & After Tom DeLay was zapped, \tokentag{Charles Colson}{PER} became {DeLay}'s personal guru.\\
\midrule
She does not sit still or lay still for you to change her \tokentag{Pampers}{PRODUCT}. & She does not sit still or lay still for you to change her \tokentag{Pampers}{PRODUCT}. & She does not sit still or lay still for you to change her \tokentag{Pampers}{PRODUCT}.\\
\midrule
Russian and Norwegian divers searched the fourth compartment of the wrecked submarine \tokentag{Kursk}{PRODUCT}, \tokentag{Sunday}{DATE}, but they found too much damage to proceed with the task of recovering bodies . & Russian and Norwegian divers searched the fourth compartment of the wrecked submarine \tokentag{Kursk}{PRODUCT}, \tokentag{Sunday}{DATE}, but they found too much damage to proceed with the task of recovering bodies . & Russian and Norwegian divers searched the fourth compartment of the wrecked submarine Kursk, \tokentag{Sunday}{DATE}, but they found too much damage to proceed with the task of recovering bodies .\\
\midrule
\tokentag{Zinni}{PER} testifying after the attack on \tokentag{the ``USS Cole''}{PRODUCT} -- Aden never had a specific terrorist threat . & \tokentag{Zinni}{PER} testifying after the attack on \tokentag{the ``USS Cole''}{PRODUCT} -- Aden never had a specific terrorist threat . & {Zinni} testifying after the attack on \tokentag{the ``USS Cole''}{PRODUCT} -- Aden never had a specific terrorist threat . \\
\midrule
Today , the enterovirus chip is in the testing phase , and DR. Chip is collaborating with Taipei Veterans General Hospital to obtain samples with which to establish the accuracy of the chip . & \tokentag{Today}{DATE}, the enterovirus chip is in the testing phase, and \tokentag{DR. Chip}{PRODUCT} is collaborating with \tokentag{Taipei Veterans General Hospital}{FAC} to obtain samples with which to establish the accuracy of the chip . & \tokentag{Today}{DATE}, the enterovirus chip is in the testing phase, and \tokentag{DR. Chip}{PRODUCT} is collaborating with {Taipei Veterans General Hospital} to obtain samples with which to establish the accuracy of the chip . \\
\midrule

\toprule
And I think perhaps no one more surprised than some of the people running those firms on \tokentag{Wall Street}{FAC}. & I think perhaps no one more surprised than some of the people running those firms on \tokentag{Wall Street}{FAC}. & I think perhaps no one more surprised than some of the people running those firms on {Wall Street}. \\
\midrule
We're all getting , this news in from the speech that the Homeland Security Secretary \tokentag{Tom Ridge}{PER} is expected to be delivering at the international press club around 1:00 Eastern at the top of the hour . & We're all getting , this news in from the speech that the Homeland Security Secretary \tokentag{Tom Ridge}{PER} is expected to be delivering at the international press club around 1:00 Eastern at the top of the hour . & We're all getting , this news in from the speech that the Homeland Security Secretary \tokentag{Tom Ridge}{PER} is expected to be delivering at the international press \tokentag{club}{FAC} around 1:00 Eastern at the top of the hour . \\
\midrule
\tokentag{Yesterday}{DATE} American pilots mechanics approved their share \tokentag{\$ 1.8 billion}{MONEY} in labor concession . & \tokentag{Yesterday}{DATE} American pilots mechanics approved their share \tokentag{\$ 1.8 billion}{MONEY} in labor concession . & {Yesterday} American pilots mechanics approved their share \$ 1.8 billion in labor concession .\\
\bottomrule

\caption{NER Prediction Examples from OntoNotes with \nertag{PERSON, DATE, MONEY, LOC, FAC,PRODUCT} as target few-shot entities} 
\label{tab:case_study}
\end{longtable}

\end{document}